\documentclass[12pt, a4paper]{article}
\usepackage[font=footnotesize]{caption}
\usepackage[utf8]{inputenc}
\usepackage{graphicx}
\usepackage{authblk}
\usepackage{booktabs}
\usepackage{verbatim}
\usepackage{url}
\usepackage{caption}
\usepackage{subcaption}
\usepackage{indentfirst}
\usepackage{url}
\usepackage{amssymb}
\usepackage{amsmath}

\usepackage{multirow}
\usepackage[table,xcdraw]{xcolor}

\usepackage{todonotes}

\usepackage{rotating}
\usepackage{array}
\title{Interrelating Fruchterman-Reingold Graph Visualization and Agglomerative Clustering}
\author{Alexandre Benatti$^1$ and \\ Luciano da F. Costa$^2$}

\affil{
$^1$Institute of Mathematics and Statistics - DCC \\
University of S\~ao Paulo \\
Rua do Mat\~ao, 1010, \\ S\~ao Paulo, SP 05508-090 Brazil 
\\ \vspace{0.5cm}
$^2$S\~ao Carlos Institute of Physics - DFCM \\
University of S\~ao Paulo \\
Av. Trabalhador S\~ao-Carlense, 400, \\ S\~ao Carlos, SP 13566-590 Brazil \\
(Prof. Senior)
}

\date{\emph{11th Sep., 2026}}

\begin{document}

\maketitle

\begin{abstract}
Graph visualization methods and agglomerative clustering have been frequently considered in data analysis and pattern recognition. Because these approaches are interrelated and complementary, it is of particular interest to investigate their associations. In this work, we study the possible relationship between the Fruchterman-Reingold graph visualization method and four types of agglomerative clustering adopting single- and complete-linkage, average, and Ward's linkage criteria. Three types of datasets have been considered in 2 and 10 dimensions, as well as the PCA projection of the latter to two dimensions. The results obtained suggest that the relationship between the methods considered did not vary much for the three types of data mentioned above. At the same time, the agglomerative methods tended to yield results that are mostly similar to each other, while presenting moderate similarity with the original data. The Fruchterman-Reingold visualization resulted similar to the original data, but exhibited relatively smaller similarity to the agglomerative methods.
\end{abstract}

\section{Introduction}

Graph visualization (e.g.~\cite{fruchterman1991, herman2000}) and agglomerative clustering (e.g.~\cite{duda2000pattern, theodoridis2006pattern}) have been frequently considered in data analysis and pattern recognition. As implied by its own name, the main objective of \emph{graph visualization} is to obtain visual representations of graphs and complex networks. At the same time, \emph{agglomerative clustering} concerns typically non-supervised methods for estimating \emph{hierarchical} relationships between data elements (characterized by respective measurements or features), which can be taken as a subsidy for searching for the existence of respective clusters. These relationships are obtained by successively merging data elements and subgroups according to some distance or similarity criterion. The results obtained by agglomerative clustering can often be represented as \emph{dendrograms}, indicating how data elements are progressively joined into subclusters.

Although graphs and dendrograms can be considered distinct structures, they are interrelated and can be used in a complementary manner. In this work, the elements of a data set (or the leaves of a respective dendrogram) will be understood to be associated to nodes in a corresponding graph. While the modular structure of a low dimensional data set can be more directly appreciated from a respective graph visualization, dendrograms provide a representation of the complete hierarchical relationships between the data elements. In addition, the latter type of representation and analysis assumes particular importance in case the data elements are characterized by more than 2 or 3 measurements (dimensions), because in these cases graph visualization, by unavoidably implementing a projection onto a two-dimensional space, cannot typically preserve the adjacency among the elements and groups of elements.

Some level of congruence can be expected between graphs and dendrograms deriving from the same source. For instance, a data set involving 4 well-separated clusters should yield a visualization characterized by 4 associated groups of nodes that are more densely interconnected to each other than to nodes of different groups. Therefore, it would be expected that a respective dendrogram obtained using agglomerative clustering would correspond to 4 main branches that are well-separated from each other in the sense of having relatively long stems. At the same time, a graph visualization in which no discernible modules (or communities) can be identified should be accompanied by a dendrogram without well-separated branches.

Interestingly, it is possible to transform (e.g.~\cite{comin2020}) between data sets, dendrograms, and graphs in several ways. For example, given a data set, the values of similarity indices or distances between the features of its elements yield a respective weighted and non-directed graph or network. At the same time, a graph can be transformed into a data set in several ways. For example, a weight matrix defining a respective graph can be obtained from the cophenetic distance (e.g.~\cite{sokal1962}) between the subgroups in a dendrogram. It is also possible to obtain comparisons between the features associated with each of the nodes in a graph, including complementary properties of nodes and/or respectively, topological measurements.

In the current work, we study a possible relationship between the often used \emph{Fruchterman-Reingold Graph Visualization} methodology and four frequently adopted types of agglomerative clustering, namely those employing single- and complete-linkage, average, and Ward's linkage criteria. As will be described, this visualization method can be more closely related, though by a little margin, to agglomerative clustering by using the Ward's criterion. The results obtained from the four considered agglomerative methods were mostly similar, while the Fruchterman-Reingold visualization presented good compatibility with the original data, but smaller similarity with the four agglomerative approaches.

This work starts by presenting a brief review of the Fruchterman-Reingold visualization method and the considered agglomerative clustering approaches, and follows by presenting and discussing the comparison between them.

\section{Fruchterman-Reingold Graph Visualization}

The visualization of graphs can be approached in several ways, of which \emph{force-based} algorithms (e.g.~\cite{fruchterman1991}) have been frequently considered. This type of approach is developed in analogy with a physical system involving several particles (the nodes) which interact with each other through attractive and repulsive forces. For example, the particles can be understood to have the same type of charge (e.g.~negative or positive), so that they repel each other. At the same time, each pair of particles can have a spring of non-zero length associated, therefore quantifying the similarity (or proximity) between pairs of nodes.

In physical terms, these two types of interactions can be associated with respective energies (potential and spring).  Given an initial configuration of nodes position, the system unfolds along time toward a minimum of energy, which may correspond to a local or global minimum. Temperature is often associated with the particle system, which controls the probability of the magnitude of state changes. Typically, the global minimum cannot be assumed to have been achieved in finite time. 

The graph visualization approach known as Fruchterman-Reingold (e.g.~\cite{fruchterman1991}) considers interactions between nodes as described above, with the electric field part of the interaction implying the particles to become more and more distributed spatially, while the springs, with respective constants related to the weights of the respective edges, tend to keep them together. In practice, the visualizations obtained by the Fruchterman-Reingold method have particles that are close to each other in the graph placed together in the respective visualization. It is also important to keep in mind that different, though congruent, graph visualizations are typically obtained by repeated application of the Fruchterman-Reingold method on the same data.

\section{Agglomerative Clustering}

Agglomerative clustering methodologies (e.g.~\cite{hill1980,rodriguez2019,tokuda2022}) can be understood aiming to progressively merge the elements in a given set of data while considering a scale variable, here represented as $s$, which can be associated with distance (e.g.~Euclidean) or similarity (e.g.~Jaccard). In the case of distance-based approaches, starting with each data element being associated with a subgroup of size 1, the pair of nearest subgroups is identified and interconnected. This basic step proceeds until a single group remains, or a maximum distance is reached. 

Several types of rules can be taken into account for merging subgroups.  In the present work, we consider the single-, complete, average, and Ward's merging criteria. Given two sets $A$ and $B$, the single-linkage criterion results in the value of the \emph{smallest} distance between all possible combinations of elements from $A$ and elements from $B$. At the same time, the complete linkage criterion takes to \emph{largest} of those distances. The average linkage yields the average of the pairwise distances between the elements taken from $A$ and $B$. The Ward's approach merges the two sets so as to minimize the resulting dispersion. 

A hierarchy of relationships between the so resulting subgroups is obtained, which is typically visualized in terms of a respective \emph{dendrogram}. Data elements which are less distant (or are more similar) tend to appear together in respective branches, the lengths of which reflect the difference between the respective elements. At the same time, the value of $s$ where a merge takes quantifies the level of interrelationship (distance or similarity) between the two sets.

\section{Methodology}

\begin{figure}[!ht]
  \centering
    \includegraphics[width=.8 \textwidth]{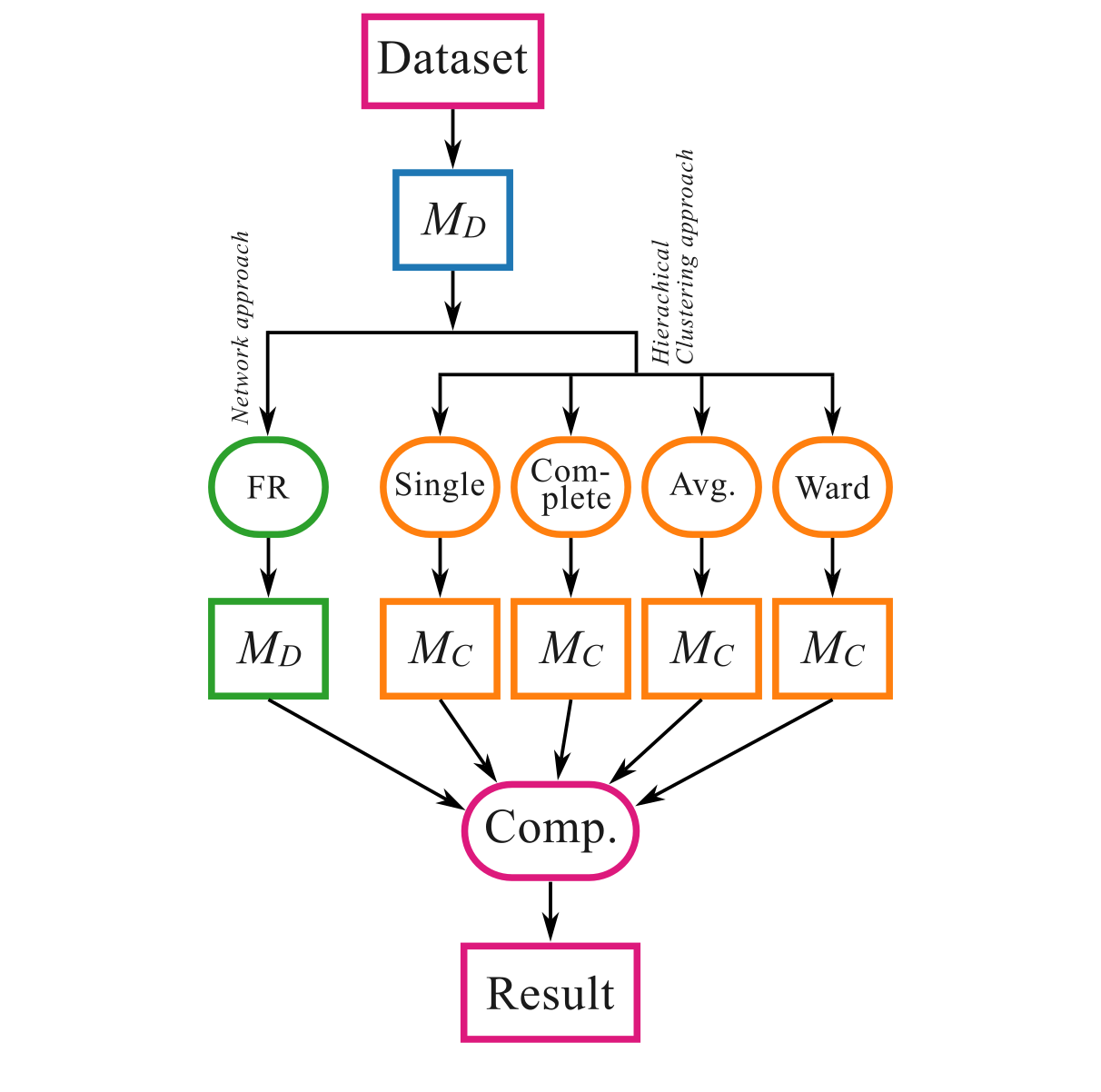}
   \caption{Flow diagram illustrating the approach adopted for relating Fructerman-Reingold graph visualization and four types of agglomerative clustering. }\label{fig:flowchart}
\end{figure}

In order to better understand possible relationships between Fruchterman-Reingold visualizations and agglomerative clustering, the methodology illustrated in the flow diagram in Figure~\ref{fig:flowchart} has been considered here.

The first step is to obtain a dataset of $N$ elements characterized by $M$ respective measurements or features. The experiments in this work consider $M=2$ and $M=10$, implying two and ten features understood to be associated with a metric space. Two-dimensional projections of the latter datasets area also taken into account. It should be observed that other configurational choices may lead to different types of results.

Because we are interested in considering relatively intricate data set structures characterized by modularity and hierarchy, synthetic data are generated hierarchically according to a respective statistical model. A set of $n_1$ points is drawn with normal probability from a $M-$dimensional space, having average corresponding to the position of the reference point and standard deviation (circularly symmetric covariance matrices are assumed). Then, each of these points is replaced by a set of $N_2$ new points drawn with normal probability $\sigma_2$, and so on.

We considered the pairwise Euclidean distance matrix of the generated data points (denoted by $M_D$) as the reference data. From this distance matrix, four different agglomerative hierarchical clustering linkage methods were evaluated: single, complete, average, and Ward. The resulting dendrograms were then converted into cophenetic distance matrices, $M_C$. In a cophenetic matrix, each entry represents the height at which two elements merge in a given dendrogram. Consequently, these cophenetic distance values depend on the linkage criterion adopted during the dendrogram construction.

Additionally, the distance matrix $M_D$ was transformed into a similarity matrix $S$ that was then used to construct a weighted similarity network. This was obtained by taking the reciprocal of the distances ($S = 1/M_D$). The Fruchterman–Reingold force-directed layout algorithm was applied to the similarity matrix to obtain a two-dimensional embedding of the data, and the pairwise Euclidean distances between the resulting positions were calculated to obtain an additional distance matrix (FR) that is used to compare with the other approaches considered. 

The comparison among these six representations was performed using only the upper triangular elements of each distance matrix, excluding the diagonal. Each vector obtained was standardized independently by subtracting its mean and dividing by its standard deviation. Finally, the standardized vectors were compared pairwise using the coincidence similarity index $C$, (e.g.~\cite{costa2022on,CostaCCompl}). This similarity index combines the Jaccard (e.g.~\cite{Jac:wiki,da2021further}) and interiority (or overlap, e.g.~\cite{vijaymeena2016a}) indices.

\section{Results and Discussion}

\begin{figure}[!ht]
  \centering
    \includegraphics[width=.99 \textwidth]{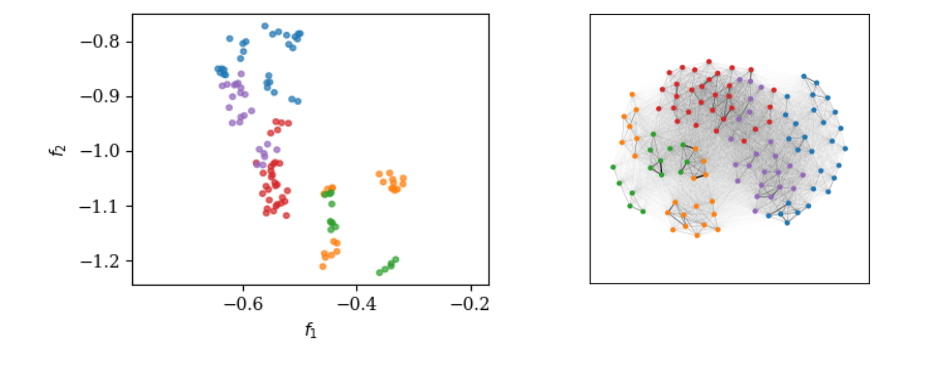}\\
    \hspace{1. cm} (a) \hspace{5. cm} (b)\\ 
   \caption{Example of 2D dataset considered in our study (a) and respective visualization by using the Fruchterman-Reingold method on the reciprocal of the Euclidean distances.}\label{fig:ex_data}
\end{figure}

We begin by presenting an example of the comparison approach respectively to the data set in Figure~\ref{fig:ex_data}(a), which is contained in a two-dimensional space. The colors indicate the groups of points which are progressively added during the adopted data generation methodology.

\begin{figure}[!ht]
  \centering
    \includegraphics[width=.8 \textwidth]{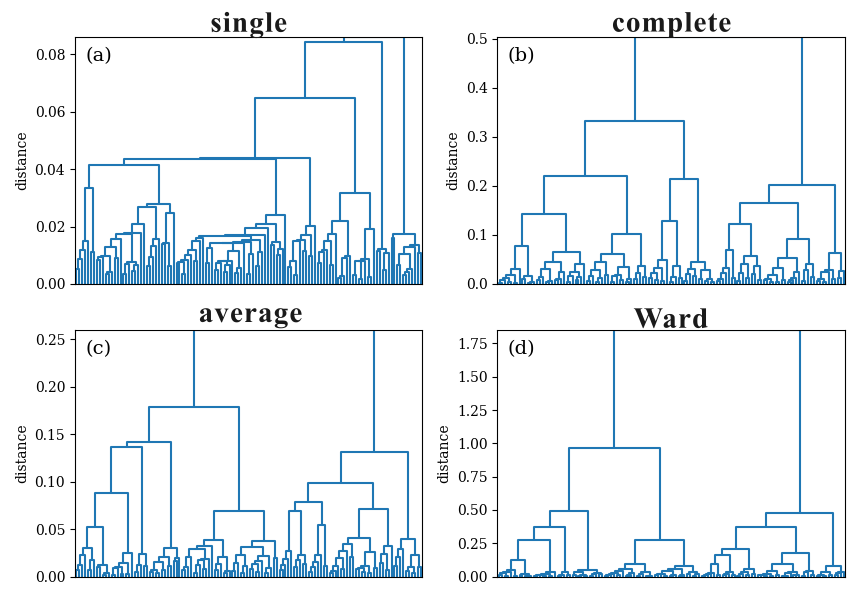} 
   \caption{Dendrograms obtained by the agglomerative clustering approaches adopting single- (a), complete- (b), average (c), and Ward's (d) linkage criteria. Observe that different upper limits of the distance axes have been chosen for the sake of improved visualization.}\label{fig:ex_dendrogram}
\end{figure}

The dendrograms obtained from the data set in Figure~\ref{fig:ex_data}(a) by using agglomerative clustering with single-, complete-, average, and Ward's linkage criteria are presented in Figure~\ref{fig:ex_dendrogram}. The structures obtained for the complete-, average, and Ward's approaches resulted mostly similar, while the dendrogram obtained by single-linkage can be observed to be more different.

\begin{figure}[!ht]
  \centering
    \includegraphics[width=.59 \textwidth]{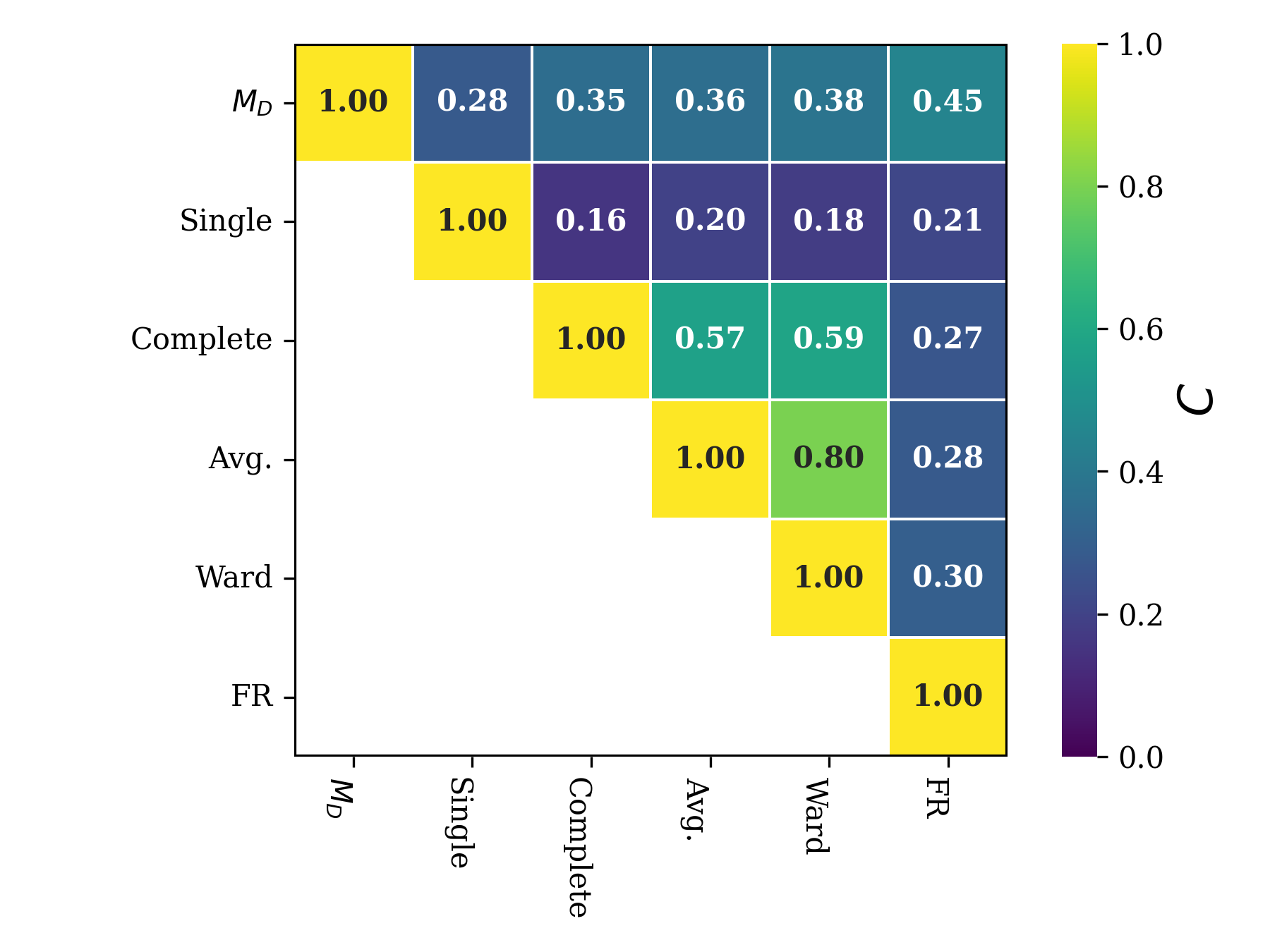}
   \caption{Coincidence similarity matrix ($D=1$) obtained for the dataset in Fig.~\ref{fig:ex_data}(a).}\label{fig:heatmap_ex}
\end{figure}

Figure~\ref{fig:heatmap_ex} shows the matrix of similarity values obtained by comparing each of the visualization and agglomerative approaches considered. In addition to confirming that the dendrogram obtained for the single-linkage methodology is less similar to those obtained by using the three other agglomerative approaches, this matrix also indicates that the original data is only moderately similar to all visualization and agglomerative methods considered in this work. At the same time, the Fruchterman-Reingold method yielded visualizations that are not particularly similar to any of the other approaches, which is discussed as follows.

Figure~\ref{fig:ex_data}(b) presents the visualization of the dataset in (a) using the Fruchterman-Reingold approach. As can be readily perceived, though the adjacency between the groups represented in colors are mostly maintained, the distances between the nodes result not only in a more compact, but also in a more uniform distribution.

The study of the relationship between the considered visualization and agglomerative approaches has been performed in terms of three experiments involving data generated as described in the previous section in two-dimensional and ten-dimensional spaces, as well as a two-dimensional projection of the latter. For each of these types of data, a total of 5,000 experiments have been performed to estimate the intrinsic relationships between the distance matrices resulting from Fruchterman-Reingold visualization and the four types of agglomerative clustering considered.

\begin{figure}[!ht]
  \centering
    \includegraphics[width=.59 \textwidth]{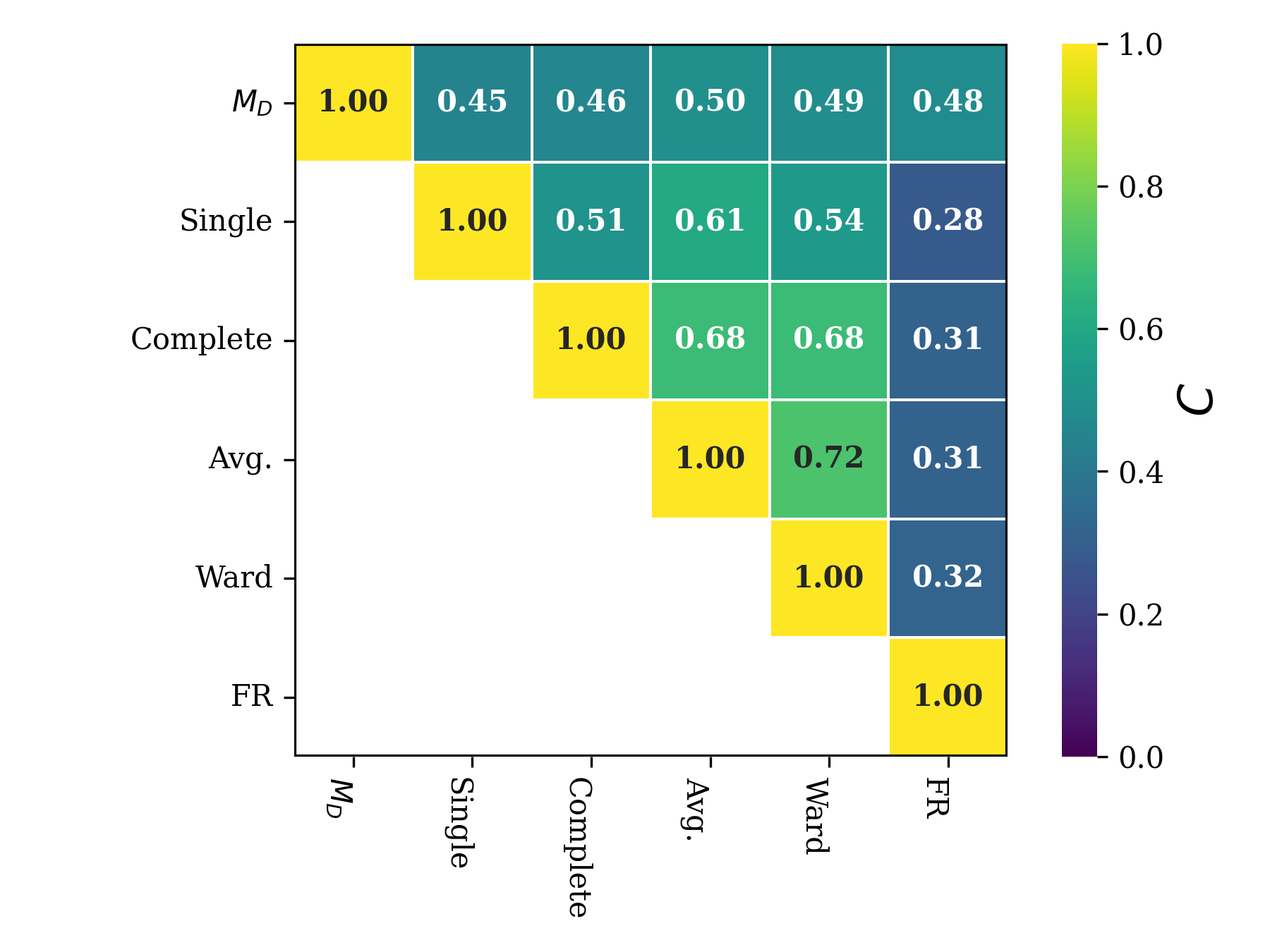}
   \caption{Average coincidence similarity matrix ($D=1$) obtained for 5,000 datasets, considering two-dimensional data.}\label{fig:heatmap}
\end{figure}

\begin{figure}[!ht]
  \centering
    \includegraphics[width=.49 \textwidth]{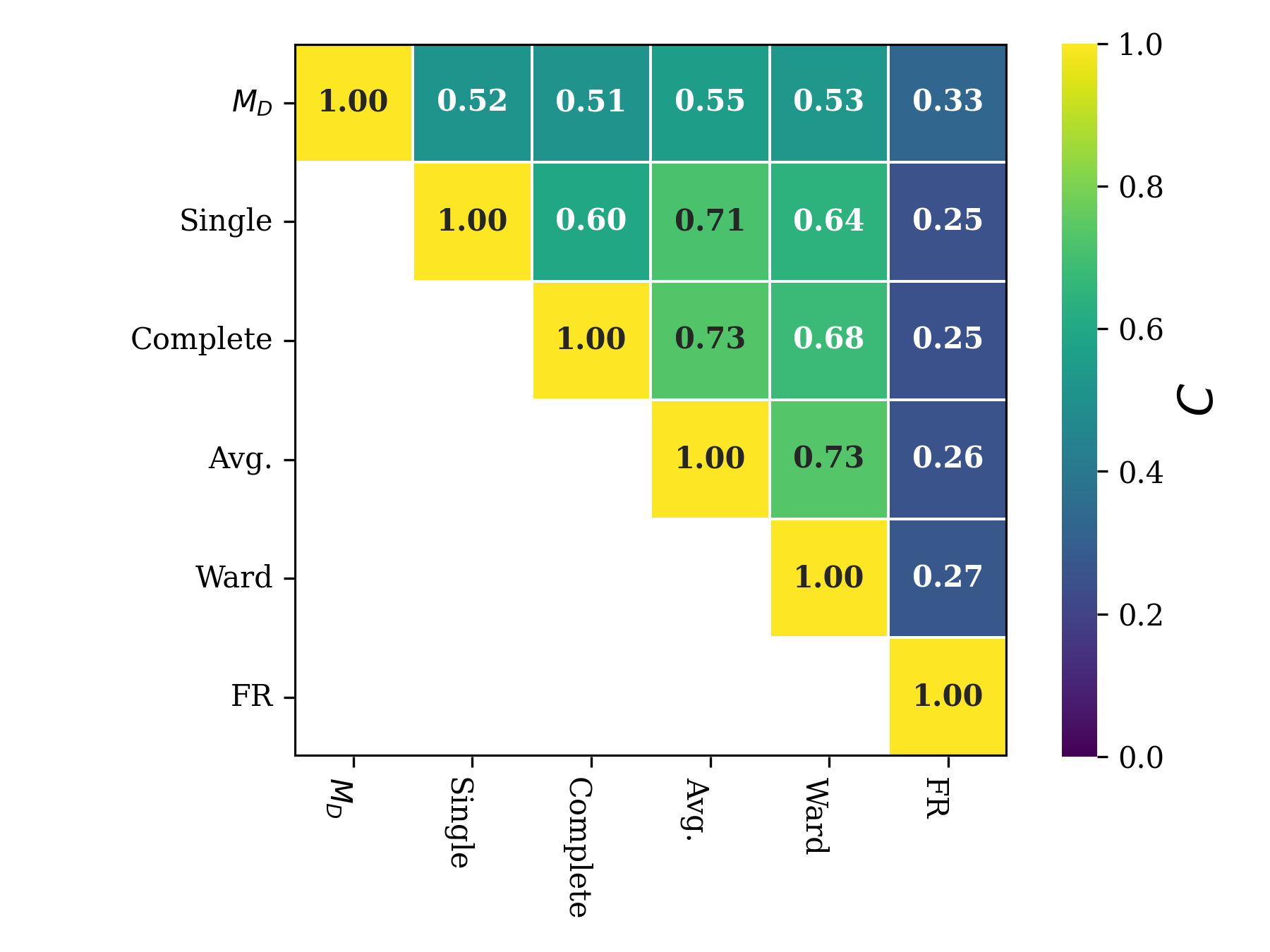}
\includegraphics[width=.49 \textwidth]{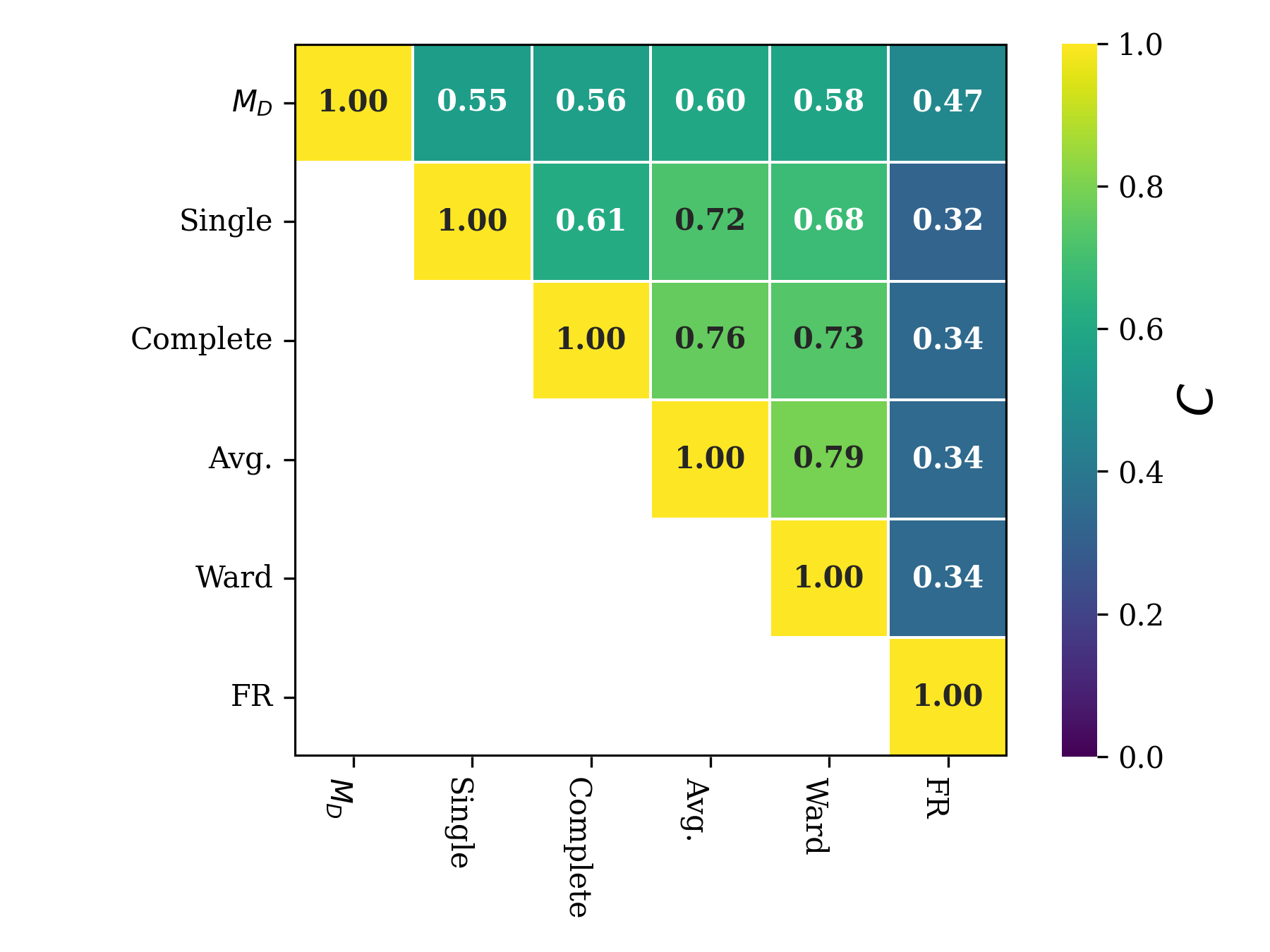}\\
\hspace{.5 cm} (a) \hspace{6.6 cm} (b)
   \caption{Average coincidence similarity matrix ($D=1$) obtained for 5,000 ten-dimensional datasets (a) and for its two-dimensional PCA projection.}\label{fig:heatmap_10d}
\end{figure}

Figures~\ref{fig:heatmap},~\ref{fig:heatmap_10d}(a), and ~\ref{fig:heatmap_10d}(b) illustrate the matrix of similarity obtained for the two- and ten-dimensional type of data, and the PCA projection of the latter, respectively.

The original dataset has been found to be reasonably similar to all agglomerative methods. Although the four agglomerative methods were found to be interrelated, smaller similarity values were obtained between the Fruchterman-Reingold visualization and the four agglomerative methods. The Fruchterman-Reingold visualization was found to be particularly similar to the original dataset in the cases of the two-dimensional type of data and two-dimensional projection of the ten-dimensional type of data. The visualization resulted less similar to the type of ten-dimensional data because of the unavoidable projection that it implements cannot preserve the original adjacency between the data elements and groups of elements.

\begin{figure}[!ht]
  \centering
    \includegraphics[width=.325 \textwidth]{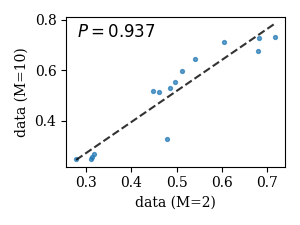}
    \includegraphics[width=.325 \textwidth]{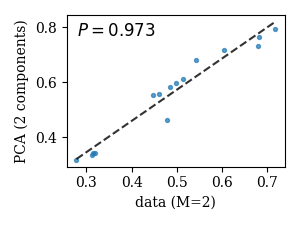}
    \includegraphics[width=.325 \textwidth]{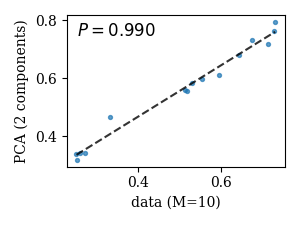}\\
    \hspace{.45 cm} (a) \hspace{3.8 cm} (b) \hspace{3.8 cm} (c)\\ 
   \caption{Correlograms relating the similarities between the visualization and agglomerative clustering obtained for three data sets considering two and ten dimensions, and the 2D PCA projection of the latter. The dashed line indicates the linear data regression.}\label{fig:correlogram}
\end{figure}

The relationship between the similarity matrix in the tables shown above is illustrated in terms of pairwise correlograms and Pearson correlation coefficients $P$ in Figure~\ref{fig:correlogram}. The results obtained indicated that these relationships are mostly compatible and congruent, meaning that the similarities between the approaches observed for the two-dimensional and ten-dimensional datasets as well as the PCA projection of the latter present similar properties.

Interestingly, the similarity relationships were in strong positive correlations, corroborating the above observations.

\section{Concluding Remarks}

Graph visualization and agglomerative clustering have often been used in several areas and from varying perspectives. Although conveying the data structure in distinct ways, some substantial level of relationship and coherence is still expected between the obtained graph visualization and dendrograms.  That is of particular interest, because these two types of representation can not only better represent the original dataset, but also provide complementary information about the original data and the problem from which they were obtained.

In the present work, an approach was developed aimed at trying to relate the frequently used Fruchterman-Reingold graph visualization methodology and agglomerative clustering based on four linkage criteria. More specifically, an experimental approach has been applied in which graph visualizations and four types of dendrograms (considering single-, complete-, average, and Ward's linkage criteria) are obtained from the same set of data elements synthesized according to a hierarchical statistical model. The visualization was then compared with the four dendrograms in terms of the distances between every pair of points. 

The obtained results suggested, for the considered data sets and respective parametric configurations, that, although the Fruchterman-Reingold graph visualization methodology tended to be more similar to the original to the two-dimensional sets of data, its similarities with data obtained the four agglomerative approaches were smaller. The results obtained using Ward's approach were found to be slightly more similar to the Fruchterman-Reingold visualization than with the dendrograms obtained for the three other linkage criteria. Interestingly, the relationships obtained were found to be mostly common to the three types of data considered.

The results obtained are of special interest because they provide subsidies for choosing among the four agglomerative clustering approaches while complementing the analysis of specific experiments and data sets by combining graph visualization and agglomerative clustering. It also provides indication, in the context of the adopted data and configurations, that the agglomerative methods preserve to a good extent the original organization of the data and that projection-based graph visualizations, such as that implemented by the Fruchterman-Reingold method, work better for lower dimensional data sets.

The approach, methodology, and results reported in the present work pave the way to several related further developments. Given that the results obtained refer to a specific model of synthesized data, other dimensions and types of data need to be taken into account in complementary experiments. It would also be of interest to consider other types of graph visualization, as well as clustering approaches.

\section*{Acknowledgments}
A. Benatti is grateful to FAPESP (grant 2025/26083-7 and 2022/15304-4). Luciano da F. Costa thanks CNPq (grant no.~313505/2023-3) and FAPESP (grant 2022/15304-4).

\bibliography{ref}
\bibliographystyle{unsrt}

\end{document}